%% file: main.tex
\documentclass{article} 
\usepackage{iclr2020_conference,times}

\input{math_commands.tex}

\usepackage{url}
\usepackage{booktabs}
\usepackage[T1]{fontenc}
\usepackage[utf8]{inputenc}
\usepackage[]{caption}
\usepackage{multirow}
\usepackage{graphicx}
\usepackage{pdfpages}
\usepackage{mathtools}
\usepackage{color}
\usepackage{soul}
\usepackage{subcaption}
\usepackage{algorithmicx}
\usepackage{nicefrac,xfrac}
\usepackage{algorithm,algpseudocode}
\usepackage{pifont}

\usepackage[toc,page]{appendix}
\usepackage{chngcntr}

\usepackage{hyperref}

\title{Consistency Regularization \\ for Generative Adversarial Networks}

\author{Han Zhang, Zizhao Zhang, Augustus Odena, Honglak Lee \\
Google Research\\
\texttt{\{zhanghan,zizhaoz,augustusodena,honglak\}@google.com} 
}

\iclrfinalcopy 
\begin{document}

\maketitle

\begin{abstract}
Generative Adversarial Networks (GANs) are known to be difficult to train, despite considerable research effort.
Several regularization techniques for stabilizing training have been proposed, but they introduce non-trivial computational overheads and interact poorly with existing techniques like spectral normalization.
In this work, we propose a simple, effective training stabilizer based on the notion of consistency regularization---a popular technique in the semi-supervised learning literature.
In particular, we augment data passing into the GAN discriminator and penalize the sensitivity of the discriminator to these augmentations. 
We conduct a series of experiments to demonstrate that consistency regularization works effectively with spectral normalization and various GAN architectures, loss functions and optimizer settings. Our method achieves the best FID scores for unconditional image generation compared to other regularization methods on CIFAR-10 and CelebA.  
Moreover, Our consistency regularized GAN (CR-GAN) improves state-of-the-art FID scores for conditional generation from 14.73 to 11.48 on  CIFAR-10 and from 8.73 to 6.66 on ImageNet-2012. 

\end{abstract}

\section{Introduction}

Generative Adversarial Networks (GANs) \citep{goodfellow2014generative} have recently demonstrated impressive results on image-synthesis benchmarks \citep{Radford15, Han17,  Miyato18b, zhang2018photographic, BIGGAN, Karras2019}. In the original setting, GANs are composed of two neural networks trained with competing goals: the \emph{generator} is trained to synthesize realistic samples to fool the discriminator and the \emph{discriminator} is trained to distinguish real samples from fake ones produced by the generator. 

One major problem with GANs is the instability of the training procedure and the general sensitivity of the results to various hyperparameters \citep{salimans2016improved}.
Because GAN training implicitly requires finding the Nash equilibrium of a non-convex game in a continuous and high dimensional parameter space, it is 
substantially more complicated than standard neural network training.
In fact, formally characterizing the convergence properties of the GAN training procedure
is mostly an open problem \citep{OPENPROBLEMS}.
Previous work \citep{ArjovskyB17, Miyato18a,odena2017conditional,ChenZRLH19, WeiGL0W18} has shown that interventions focused on the discriminator can mitigate stability issues. 
Most successful interventions fall into two categories, normalization and regularization.
Spectral normalization is the most effective normalization method, in which weight matrices in the discriminator are divided by an approximation of their largest singular value.
For regularization, \citet{WGANGP} penalize the gradient norm of straight lines between real data and generated data.
\citet{RothLNH17} propose to directly regularize the squared gradient norm for both the training data and the generated data.
DRAGAN \citep{kodali2017convergence} introduces another form of gradient penalty where the gradients at Gaussian perturbations of training data are penalized. One may anticipate simultaneous regularization and normalization could improve sample quality.  
However, most of these gradient based regularization methods either provide marginal gains or fail to introduce any improvement when normalization is used \citep{compare_gan}, which is also observed in our experiments. These regularization methods and spectral normalization are motivated by controlling Lipschitz constant of the discriminator. We suspect this might be the reason that applying both does not lead to overlaid gain.

In this paper, we examine a technique called consistency regularization \citep{BachmanAP14, CONSISTENCY,laine2016temporal, CONSISTENCYAVITAL,UDACONSISTENCY, HuMTMS17} in contrast to gradient-based regularizers. Consistency regularization is widely used in semi-supervised learning to ensure that the classifier output remains unaffected for an unlabeled example even it is augmented in semantic-preserving ways. In light of this intuition, we hypothesize a well-trained discriminator should also be regularized to have the consistency property, which enforces the discriminator to be unchanged by arbitrary semantic-preserving perturbations and to focus more on semantic and structural changes between real and fake data. Therefore, we propose a simple regularizer to the discriminator of GAN: we augment images with semantic-preserving augmentations before they are fed into the GAN discriminator and penalize the sensitivity of the discriminator to those augmentations.

\begin{figure}[t]
    \centering
    \includegraphics[width=140mm]{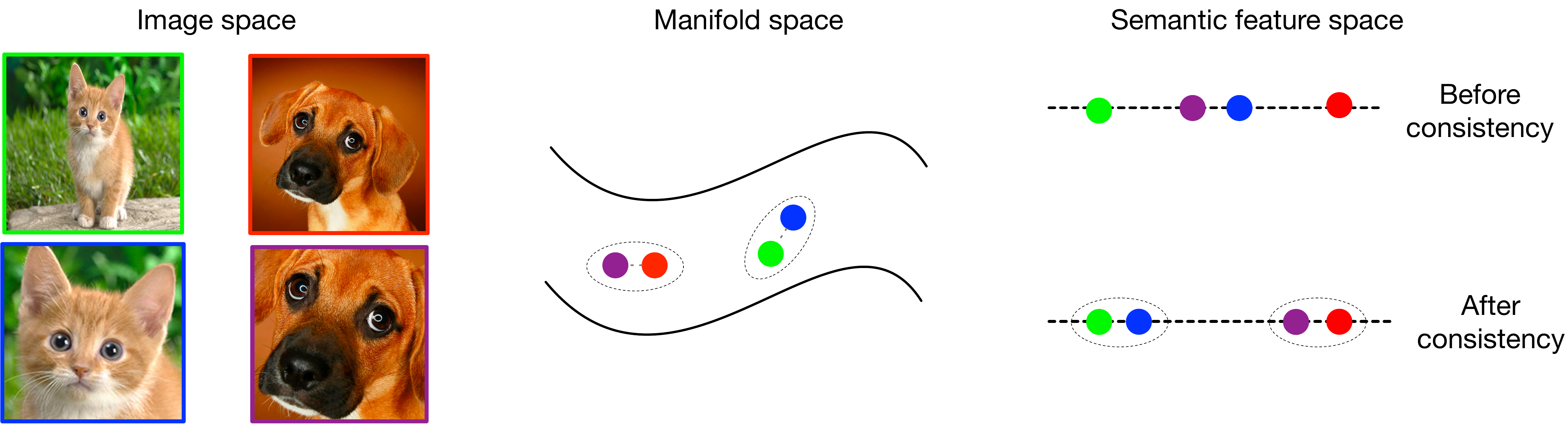}
    \caption{
    An illustration of consistency regularization for GANs. Before consistency regularization, 
    the zoomed-in dog and the zoomed-in cat (bottom left) can be closer than they are to their original images in feature space induced by the GAN discriminator.
    This is illustrated in the upper right (the semantic feature space), where the purple dot is closer to the 
    blue dot than to the red dot, and so forth.
    After we enforce consistency regularization based on the implicit assumption 
    that image augmentation preserves the semantics we care about, the purple dot pulled
    closer to the red dot.
    }
    \label{fig:illustration}
\end{figure}

This technique is simple to use and surprisingly effective.
It is as well less computationally expensive than prior techniques. 
More importantly, in our experiments, consistency regularization can always further improve the model performance when spectral normalization is used, whereas the performance gains of previous regularization methods diminish in such case. In extensive ablation studies, we show that it works across a large range of GAN variants
and datasets.
We also show that simply applying this technique on top of existing GAN models leads to
new state-of-the-art results as measured by Frechet Inception Distance \citep{FID}.

In summary, our contributions are summarized as follows:
    \begin{itemize}
    \item We propose consistency regularization for GAN discriminators to yield a simple, effective regularizer with lower computational cost than gradient-based regularization methods.
    \item We conduct extensive experiments with different GAN variants to demonstrate that our technique interacts effectively with spectral normalization. Our consistency regularized GAN (CR-GAN) achieves the best FID scores for unconditional image generation on both CIFAR-10 and CelebA.
    \item  We show that simply applying the proposed technique can further boost the performance of state-of-the-art GAN models. We improve FID scores for conditional image generation from 14.73 to 11.48 on CIFAR-10 and from 8.73 to 6.66 on ImageNet-2012.
\end{itemize}

\section{Method}

\subsection{GANs} \label{sec:gans}
A GAN consists of a generator network and a discriminator network.
The generator $G$ takes a latent variable $z \sim p{(z)}$ sampled from a prior distribution and maps it to the observation space $\mathcal{X}$.
The discriminator $D$ takes an observation $x \in \mathcal{X}$ and produces a decision output over possible observation sources (either from $G$ or from the empirical data distribution). 
In the standard GAN training procedure the generator $G$ and the discriminator $D$ are trained by minimizing the following objectives in an alternating fashion:
\begin{equation}
\begin{split}
L_D&=-\mathbb{E}_{x \sim p_\text{data}}\left[\log D(x)\right] - \mathbb{E}_{z \sim p(z)}\left[1 - \log D(G(z))\right], \\
L_G&=- \mathbb{E}_{z \sim p(z)}\left[\log D(G(z))\right], 
\end{split} \label{eq:ns_gan}
\end{equation}
where $p(z)$ is usually a standard normal distribution. This formulation is originally proposed by \citet{goodfellow2014generative} as non-saturating (NS) GAN. A significant amount of research has been done on modifying this formulation in order to improve the training process. A notable example is the hinge-loss version of the adversarial loss~\citep{lim2017, Tran2017}:
\begin{equation}
\begin{split}
L_D&=-\mathbb{E}_{x \sim p_\text{data}}\left[\min(0, -1+D(x))\right] - \mathbb{E}_{z \sim p(z)}\left[\min(0, -1-D(G(z)))\right], \\
L_G&=- \mathbb{E}_{z \sim p(z)}\left[D(G(z))\right]. 
\end{split} \label{eq:hinge_gan}
\end{equation}
Another commonly adopted GAN formulation is the Wassertein GAN (WGAN) \citep{WGAN}, in which the authors propose clipping the weights of the discriminator in an attempt to enforce that the GAN training procedure implicitly optimizes
a bound on the Wassertein distance between the target distribution and the distribution given by the generator. The loss function of WGAN can be written as
\begin{equation}
\begin{split}
L_D&=-\mathbb{E}_{x \sim p_\text{data}}\left[D(x)\right] + \mathbb{E}_{z \sim p(z)}\left[D(G(z))\right], \\
L_G&=- \mathbb{E}_{z \sim p(z)}\left[D(G(z))\right].
\end{split} \label{eq:wass_gan}
\end{equation}
Subsequent work has refined this technique in several ways 
\citep{WGANGP,Miyato18a,SAGAN},
and the current widely-used practice is to enforce spectral normalization~\citep{Miyato18a} on both the generator and the discriminator.

\subsection{Consistency Regularization}
Consistency regularization has emerged as a gold-standard technique \citep{CONSISTENCY,laine2016temporal,CONSISTENCYAVITAL,UDACONSISTENCY,REALISTICSSL, mixmatch2019} for
semi-supervised learning on image data.
The basic idea is simple: an input image is perturbed in some semantics-preserving ways
and the sensitivity of the classifier to that perturbation is penalized. 
The perturbation can take many forms: it can be image flipping, or cropping, or adversarial attacks.
The regularization form is either the mean-squared-error \citep{CONSISTENCY,laine2016temporal} between the model's output for a perturbed and non-perturbed input or the KL divergence \citep{UDACONSISTENCY, miyato2018virtual} between the distribution over classes implied by the output logits.

\subsection{Consistency Regularization for GANs}
The goal of the discriminator in GANs is to distinguish real data from fake ones produced by the generator. 
The decision should be invariant to any valid domain-specific data augmentations. For example, in the image domain, the image being real or not should not change if we flip the image horizontally or translate the image by a few pixels. However, the discriminator in GANs does not guarantee this property explicitly. 

To resolve this, we propose a consistency regularization on the GAN discriminator during
training.
In practice, we randomly augment training images as they are passed to 
the discriminator and penalize the sensitivity of the discriminator to those augmentations.

We use $D_j(x)$ to denote the output vector before activation of the $j$th layer of the discriminator
given input $x$. 
$T(x)$ denotes a stochastic data augmentation function. 
This function can be linear or nonlinear, but aims to preserve the 
semantics of the input. 
Our proposed regularization is given by
\begin{equation}
\min_{D} \; L_{cr} = \;  \min_{D} \sum_{j=m}^{n} 
  \lambda_j \big\lVert D_j(x) - D_j(T(x)) \big\rVert^2,
\label{eq:ori_eq}
\end{equation}
where $j$ indexes the layers, $m$ is the starting layer and $n$ is the ending layer that consistency is enforced.
$\lambda_j$ is weight coefficient for $j$th layer and  $\lVert \cdot \rVert$ denotes $\normltwo$ norm of a given vector. 
This consistency regularization encourages the discriminator to produce the same output for a data point under various data augmentations.

In our experiments, we find that consistency regularization on the last layer of the discriminator before the activation function is sufficient. 
$L_{cr}$ can be rewritten as
\begin{equation}
L_{cr} = \big\lVert D(x) - D(T(x)) \big\rVert^2,
\label{eq:ori_simple}
\end{equation}
where from now on we will drop the layer index for brevity. 
This cost is added to the discriminator loss (weighted by a hyper-parameter $\lambda$) when updating the discriminator parameters.
The generator update remains unchanged.
Thus, the overall consistency regularized GAN (CR-GAN) objective is written as
\begin{equation}
L_D^{cr}= L_D + \lambda L_{cr},\qquad  L_G^{cr} = L_G.
\label{eq:ns_gan_final}
\end{equation}
Our design of $L_{cr}$ is general-purpose and thereby can work with any valid adversarial losses $L_G$ and $L_D$ for GANs (See Section \ref{sec:gans} for examples). 
Algorithm \ref{alg:main} illustrates the details of CR-GAN with Wassertein loss as an example. In contrast to previous regularizers, our method does not increase much overhead. The only extra computational cost comes from feeding an additional (third) image through the discriminator forward and backward when updating the discriminator
parameters.

\begin{algorithm}[t]
    \caption{Consistency Regularized GAN (CR-GAN). We use $\lambda=10$ by default.} 
    \label{alg:main}
\begin{algorithmic}[1]
    \renewcommand{\algorithmicrequire}{\textbf{Input:}}
    \renewcommand{\algorithmicensure}{\textbf{Output:}}
    \Require generator and discriminator parameters $\theta_G, \theta_D$, consistency regularization coefficient $\lambda$, Adam hyperparameters $\alpha, \beta_1, \beta_2$, batch size $M$, number of discriminator iterations per generator iteration $N_D$
    
    \For{number of training iterations} 
    \For{$t=1,...,N_D$}            
    \For{$i=1,...,M$}
    \State Sample $z \sim p(z)$,  $x \sim p_\text{data}(x)$
    \State Augment $x$ to get $T(x)$
    \State $L_{cr}^{(i)} \gets \big\lVert D(x) - D(T(x)) \big\rVert^2$
    \State $L_D^{(i)} \gets D(G(z))-D(x)$
    \EndFor
    \State $\theta_D \gets \text{Adam}(\frac{1}{M} \sum_{i=1}^M ( L_D^{(i)}+ \lambda L_{cr}^{(i)}), \alpha, \beta_1, \beta_2)$
    \EndFor
    \State  Sample a batch of latent variables $\{z^{(i)}\}_{i=1}^{M} \sim p(z)$  
    \State $\theta_G \gets \text{Adam}(\frac{1}{M} \sum_{i=1}^M (-D(G(z))), \alpha, \beta_1, \beta_2)$
    \EndFor
\end{algorithmic}
\end{algorithm}

\section{Experiments}
\label{section:experiments}

This section validates our proposed CR-GAN method. First we conduct a large scale study to compare consistency regularization to existing GAN regularization techniques \citep{kodali2017convergence, WGANGP, RothLNH17} for several GAN architectures, loss functions and other hyper-parameter settings. 
We then apply consistency regularization to a state-of-the-art GAN model \citep{BIGGAN} 
and demonstrate performance improvement.
Finally, we conduct ablation studies to investigate the importance of various design choices 
and hyper-parameters. 
All our experiments are based on the open-source code from Compare GAN \citep{compare_gan},
which is available at 
{\href{https://github.com/google/compare\_gan}{https://github.com/google/compare\_gan.}}

\subsection{Datasets and Evaluation Metrics}
We validate our proposed method on three datasets: CIFAR-10 \citep{cifar10}, 
CELEBA-HQ-128 \citep{PROGRESSIVEGAN}, and ImageNet-2012 \citep{IMAGENET}.
We follow the procedure in \citet{compare_gan} to prepare datasets.
CIFAR-10 consists of 60K of $32 \times 32$ images in 10 classes; 
50K for training and 10K for testing.
CELEBA-HQ-128 (CelebA) contains 30K images of faces at a resolution of $128 \times 128$. 
We use 3K images for testing and the rest of images for training.
ImageNet-2012 contains roughly 1.2 million images with 1000 distinct categories and we down-sample the images to $128 \times 128$ in our experiments.

We adopt the Fr\'echet Inception distance (FID)~\citep{FID} as primitive metric for quantitative evaluation, 
as FID has proved be more consistent with human evaluation. 
In our experiments the FID is calculated on the test dataset.
In particular, we use 10K generated images vs. 10K test images on CIFAR-10, 3K vs. 3K on CelebA and 50K vs. 50K on ImageNet. We also provide the Inception Score~\citep{salimans2016improved} for different methods in the Appendix \ref{sec:is_section} for supplementary results. 
By default, the augmentation used in consistency regularization is a combination of randomly 
shifting the image by a few pixels and randomly flipping the image horizontally.
The shift size is 4 pixels for CIFAR-10 and CelebA and 16 for ImageNet.

\subsection{Comparison with other GAN regularization methods}
In this section, we compare our methods with three GAN regularization techniques,  Gradient Penalty (GP) \citep{WGANGP}, DRAGAN Regularizer (DR) \citep{kodali2017convergence} and JS-Regularizer (JSR) \citep{RothLNH17} on CIFAR-10 and CelebA. 

Following the procedures from \citep{compare_gan, LucicKMGB18}, we evaluate these methods across different optimizer parameters, loss functions, regularization coefficient and neural architectures.
For optimization, we use the Adam optimizer with batch size of 64 for all our experiments.
We stop training after 200k generator update steps for CIFAR-10 and 100k steps for CelebA. By default, spectral normalization (SN) \citep{Miyato18a} is used in the discriminator, 
as this is the most effective normalization method for GANs \citep{compare_gan} and is becoming the standard for `modern' GANs \citep{SAGAN,BIGGAN}.
Results without spectral normalization can be seen in the Appendix \ref{sec:no_sn}.

\subsubsection{Impact of Loss function} \label{sec: impact_loss}

\begin{figure}[t]
    \centering
    \includegraphics[width=130mm]{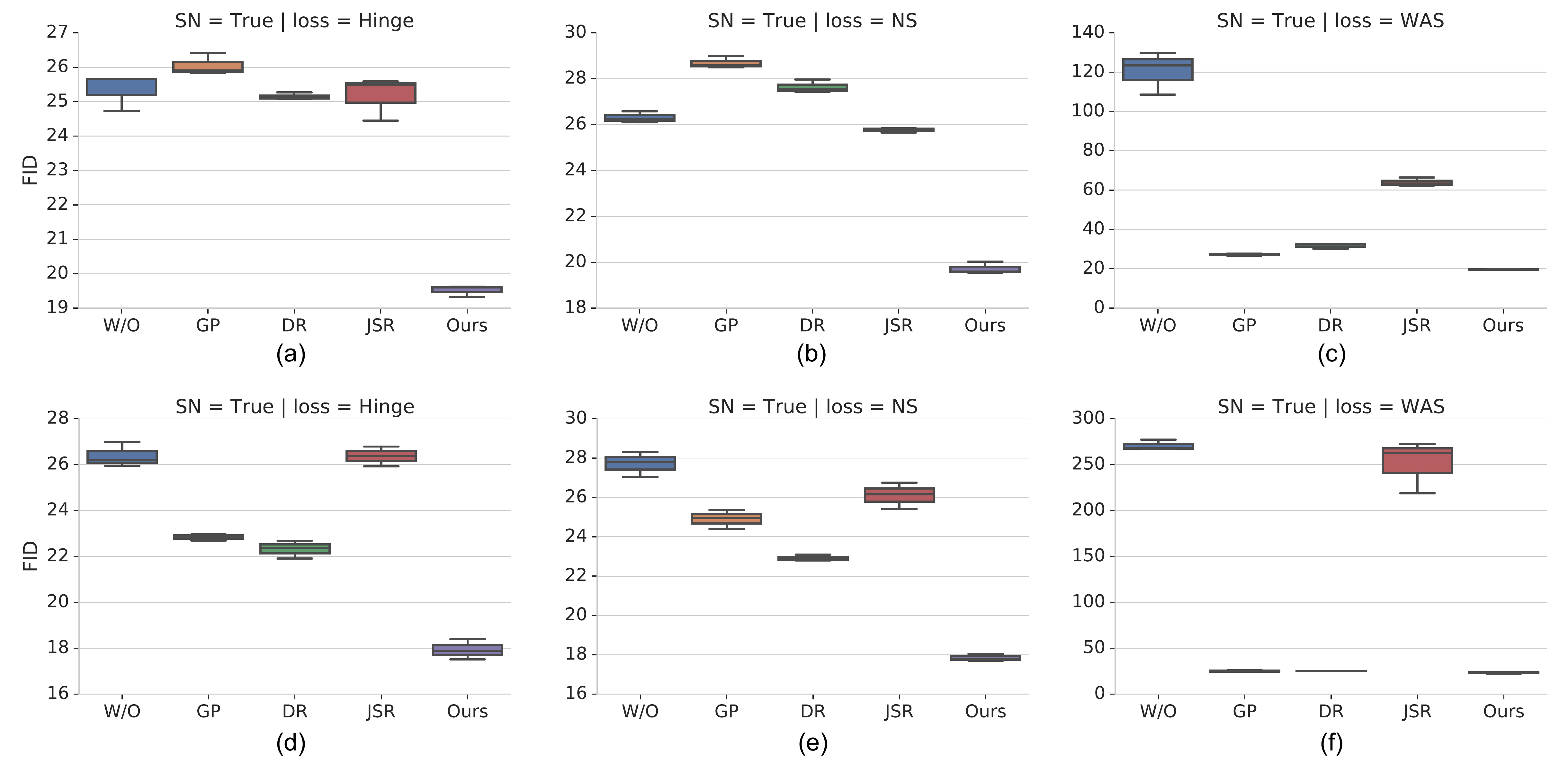}
    \caption{
    Comparison of our method with existing regularization techniques under different GAN losses.
    Techniques include no regularization (W/O), Gradient Penalty (GP) \citep{WGANGP}, DRAGAN (DR) \citep{kodali2017convergence} and JS-Regularizer (JSR) \citep{RothLNH17}.
    Results (a-c) are for CIFAR-10 and results (d-f) are for CelebA.  
    }
    \label{fig:loss_fn}
\end{figure}
In this section, we discuss how each regularization method performs when the loss function is changed. 
Specifically, we evaluate regularization methods using three loss functions: 
the non-saturating loss (NS) \citep{goodfellow2014generative},
the Wasserstein loss (WAS) \citep{WGAN}, and 
the hinge loss (Hinge) \citep{lim2017, Tran2017}.
For each loss function, we evaluate over 7 hyper-parameter settings of the Adam optimizer 
(more details in Section \ref{sec:hyper_settings} of the appendix). 
For each configuration, we run each model 3 times with different random seeds.
For the regularization coefficient, we use the best value reported in the corresponding paper.
Specifically $\lambda$ is set to be 10 for both GP, DR and our method and 0.1 for JSR.
In this experiment, we use the SNDCGAN network architecture \citep{Miyato18a} for simplicity.
In the end, similar as \citet{compare_gan},  we aggregate all runs and report the FID distribution of the top 15\% of trained models.

The results are shown in Figure~\ref{fig:loss_fn}. The consistency regularization improves the baseline across all different loss functions and both datasets. 
Other techniques have more mixed results:
For example, GP and DR can marginally improve the performance for settings (d) and (e) but lead to worse results for settings (a) and (b) (which is consistent with findings from \citet{compare_gan}). 
In all cases, our consistency-regularized GAN models have the lowest (best) FID. 

This finding is especially encouraging, considering that the consistency regularization has lower computational cost (and is simpler to implement) than the other techniques. In our experiments, the consistency regularization is around $1.7$ times faster than gradient based regularization techniques, including DR, GP and JSR, which need to compute the gradient of the gradient norm $\lVert \nabla_x(D) \rVert$. Please see Table \ref{tab:speed_comparison} in the appendix for the actual training speed.

\begin{table}[t]
\centering
\begin{tabular}{l|ccccc}
 \hline
Setting   & W/O &  GP & DR & JSR  & Ours (CR-GAN) \\ 
\hline
 CIFAR-10 (SNDCGAN)  & 24.73 & 25.83  & 25.08  & 25.17   & \textbf{18.72} \\
  CIFAR-10 (ResNet)  &19.00 & 19.74  & 18.94  & 19.59 &   \textbf{14.56}\\ 
 \hline
 CelebA (SNDCGAN)  & 25.95 & 22.57  & 21.91 &  22.17 & \textbf{16.97}\\

 \hline \hline
\end{tabular}
\caption{Best FID scores for unconditional image generation on CIFAR-10 and CelebA.
} 
\label{tab:main_comparison}
\end{table}

\subsubsection{Impact of the regularization coefficient}

\begin{figure}[t]
    \centering
    \includegraphics[width=140mm]{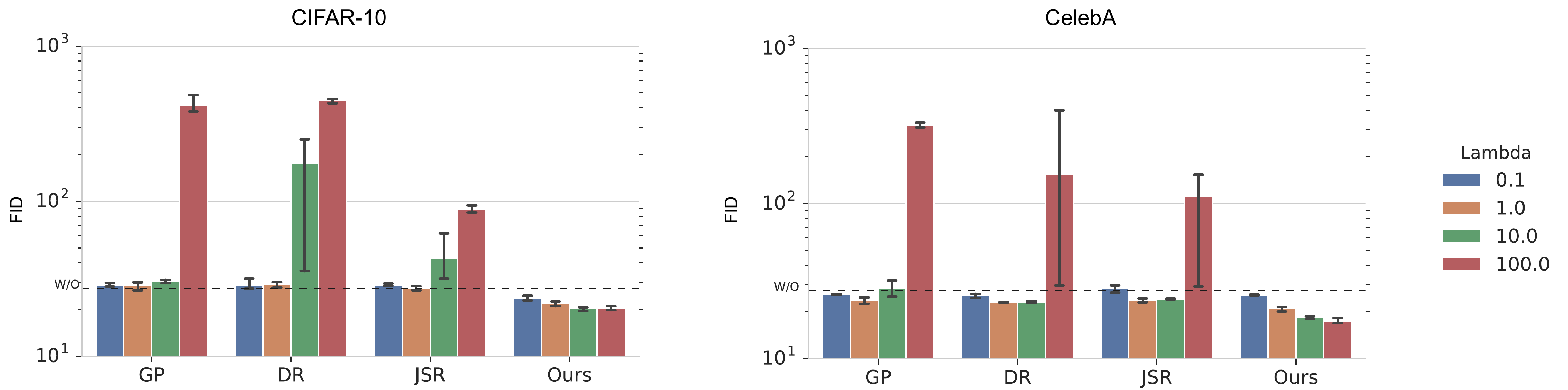}
    \caption{
    Comparison of FID scores with different values of the regularization coefficient $\lambda$ 
    on CIFAR-10 and CelebA. The dotted line is a model without regularization.
    }
    \label{fig:lambda_sweep}
\end{figure}

Here we study the sensitivity of GAN regularization techniques to the regularization coefficient $\lambda$.
We train SNDCGANs with non-saturating losses and fix the other hyper-parameters.
$\lambda$ is chosen among \{0.1, 1, 10, 100\}.
The results are shown in Figure \ref{fig:lambda_sweep}.
From this figure, we can see consistency regularization is more robust to changes in $\lambda$ than other
GAN regularization techniques (it also has the best FID for both datasets).
The results indicate that consistency regularization can be used as a plug-and-play technique
to improve GAN performance in different settings without much hyper-parameter tuning.

\subsubsection{Impact of Neural Architectures}

\begin{figure}[t]
    \centering
    \includegraphics[width=120mm]{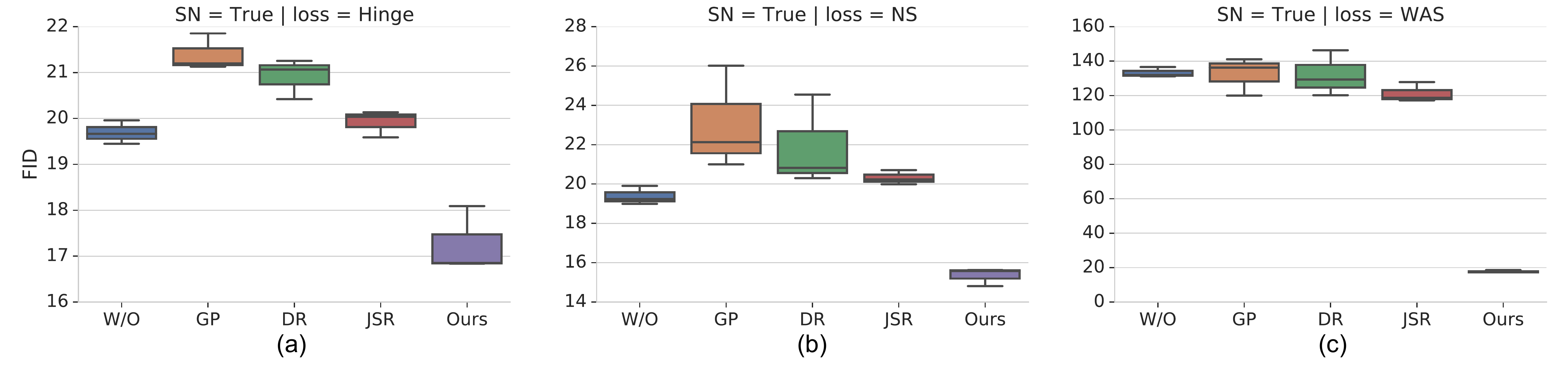}
    \caption{
    Comparison of FID scores with ResNet structure on different loss settings on CIFAR-10. 
    }
    \label{fig:resnet_loss}
\end{figure}

To validate whether the above findings hold across different neural architectures, 
we conduct experiments on CIFAR-10 using a ResNet \citep{HeZRS15, WGANGP} architecture instead of an 
SNDCGAN.
All other experimental settings are same as in Section \ref{sec: impact_loss}.
The FID values are presented in Figure \ref{fig:resnet_loss}.
By comparing results in Figure \ref{fig:resnet_loss} and Figure \ref{fig:loss_fn}, 
we can see that results on SNDCGAN and results on ResNet are comparable, though 
consistency regularization favors even better in this case:
In sub-plot (c) of Figure~\ref{fig:resnet_loss}, we can see that consistency regularization is the only regularization method that can generate satisfactory samples with a reasonable FID score (The FID scores for other methods are above 100). 
Please see Figure \ref{fig:stability_sample} for the actual generated samples in this setting. 
As in Section \ref{sec: impact_loss}, consistency regularization has the best FID for each setting.

In Table \ref{tab:main_comparison}, we show FID scores for the best-case settings from this section.
Consistency regularization improves on the baseline by a large margin and achieves the best results across different network architectures and datasets.
In particular, it achieves an FID 14.56 on CIFAR-10 16.97 on CelebA. 
In fact, our FID score of 14.56 on CIFAR-10 for \textit{unconditional} image generation is even lower than the 14.73 reported in \citet{BIGGAN} for
\textit{class-conditional} image-synthesis with a much larger network architecture and much bigger batch size.


\subsection{Comparison with state-of-the-art GAN models}

In this section, we add consistency regularization to the state-of-the-art BigGAN
model \citep{BIGGAN} and perform class conditional image-synthesis on CIFAR-10 and ImageNet.
Our model has exactly the same architecture and is trained under the same settings as BigGAN$^\star$, the open-source implementation of BigGAN from \citet{compare_gan}.
The only difference is that our model uses consistency regularization. 
In Table~\ref{tab:compare_others}, we report the original FID scores without noise truncation.
Consistency regularization improves the FID score of BigGAN$^\star$ on CIFAR-10 from 20.42 to 11.48.
In addition, the FID on ImageNet is improved from 7.75 to 6.66. 

Generated samples for CIFAR-10 and ImageNet with consistency regularized models and baseline models are shown in Figures \ref{fig:cifar_samples_conditional}, \ref{fig:imagenet_ours} and \ref{fig:imagenet_baseline} in the appendix.

\begin{table*}[hbt]
\begin{center}
\begin{tabular}{l|ccccc}
\hline
Dataset  & SNGAN &SAGAN  & BigGAN  & BigGAN$^\star$ & CR-BigGAN$^\star$ \\
\hline
CIFAR-10 & 17.5  & /&  14.73  &20.42 & \textbf{11.48} \\
ImageNet & 27.62 & 18.65 &  8.73    & 7.75 & \textbf{6.66} \\ 

\hline
\hline
\end{tabular}
\end{center}
\vspace{-6pt}
    \caption{Comparison of our technique with state-of-the-art GAN models including SNGAN \citep{Miyato18b}, SAGAN \citep{SAGAN} and BigGAN \citep{BIGGAN} 
    for class conditional image generation on CIFAR-10 and ImageNet in terms of FID.
    BigGAN$^\star$ is the BigGAN implementation of \citet{compare_gan}.
    CR-BigGAN$^\star$ has the exactly same architecture as BigGAN$^\star$ and is trained with the same settings.
    The only difference is CR-BigGAN$^\star$ adds consistency regularization.}
\label{tab:compare_others} 
\end{table*}
%
%

\section{Ablation Studies and Discussion}

\subsection{How Much Does Augmentation Matter by Itself?}

\begin{figure}[t]
    \centering
    \includegraphics[width=130mm]{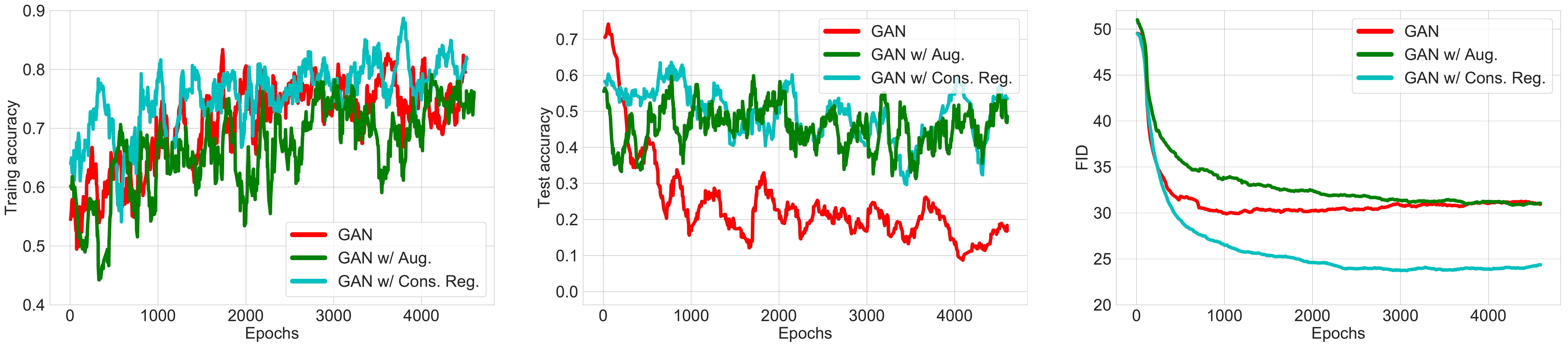}
    \caption{
    A study of how much data augmentation matters by itself.
    Three GANs were trained on CIFAR-10: one baseline GAN, one GAN with data augmentation 
    only, and one GAN with consistency regularization.
    (\textbf{Left}) Training accuracy of the GAN discriminator. 
    (\textbf{Middle}) Test accuracy of the GAN discriminator on the held out test set. 
    The accuracy is low for the baseline GAN, which indicates it suffered from over-fitting. 
    The accuracy for the other two is
    basically indistinguishable for each other. This suggests that augmentation by itself is enough to reduce discriminator
    over-fitting, and that consistency regularization by itself does little to address
    over-fitting.
    (\textbf{Right})
    FID scores of the three settings. 
    The score for the GAN with only augmentation is not any better than the score for the
    baseline, even though its discriminator is not over-fitting.
    The score for the GAN with consistency regularization is better than both of the others, suggesting that the consistency regularization acts on the score through some 
    mechanism other than by reducing discriminator over-fitting.
    }
    \label{fig:d_loss}
\end{figure}

Our consistency regularization technique actually has two parts: 
we perform data augmentation on inputs from the training data,
and then consistency is enforced between the augmented data and the 
original data. We are interested in whether the performance gains shown in Section \ref{section:experiments} are merely due to data augmentation, since data augmentation reduces the over-fitting of the discriminator to the input data. Therefore, we have designed an experiment to answer this question. 
First, we train three GANs:
(1) a GAN trained with consistency regularization, as in Algorithm \ref{alg:main},
(2) a baseline GAN trained without augmentation or consistency regularization, and 
(3) a GAN trained with only data augmentation and no consistency regularization.
We then plot (Figure~\ref{fig:d_loss}) both their FID and the test accuracy of their discriminator on a held-out 
test set.
The FID tells us how `good' the resulting GAN is, and the discriminator test accuracy tells us
how much the GAN discriminator over-fits. Interestingly, we find that these two measures are not well correlated in this case. 
The model trained with only data augmentation over-fits substantially less than the baseline
GAN, but has almost the same FID. 
The model trained with consistency regularization has the same amount of over-fitting 
as the model trained with just data augmentation, but a much lower FID. 

This suggests an interesting hypothesis, which is that the mechanism by which the consistency regularization improves GANs is not simply discriminator generalization (in terms of classifying images into real vs fake). We believe that the main reason for the impressive gain from the consistency regularization is due to learning more semantically meaningful representation for the discriminator. More specifically, data augmentation will simply treat all real images and their transformed images with the same label as real without considering semantics, whereas our consistency regularization further enforces learning implicit manifold structure in the discriminator that pulls semantically similar images (i.e., original real image and the transformed image) to be closer in the discriminator representation space.

\subsection{How does the Type of Augmentation Affect Results?}
To analyze how different types of data augmentation affect our results, we conduct an ablation study on the CIFAR-10 dataset comparing the results of using four different types of image augmentation: 
(1) adding Gaussian noise to the image in pixel-space,
(2) randomly shifting the image by a few pixels and randomly flipping it horizontally,
(3) applying cutout \citep{devries2017improved} transformations to the image, and
(4) cutout \textit{and} random shifting and flipping. 
\begin{table}[t]
\centering
\begin{tabular}{l|cccc}
 \hline
Metric   & Gaussian Noise &  Random shift \& flip & Cutout & Cutout w/ random shift \& flip \\ \cline{1-5}
 FID     & 21.91$\pm0.32$   & 16.04$\pm0.17$  &   17.10$\pm0.29$  &   19.46$\pm0.26$    \\
 \hline \hline
\end{tabular}
\caption{
FID scores on CIFAR-10 for different types of image augmentation.
Gaussian noise is the worst, and random shift and flip is the best, consistent with 
general consensus on the best way to perform image optimization on CIFAR-10 \citep{Zagoruyko2016WRN}.
} 
\label{tab:different_aug}
\end{table}
As shown in Table \ref{tab:different_aug}, random flipping and shifting \textit{without} cutout gives the best results (FID 16.04) among all four methods.
Adding Gaussian noise in pixel-space gives the worst results.
This result empirically suggests that adding Gaussian noise is not a good semantic preserving transformation in the image manifold.  
It's also noteworthy that the most extensive augmentation (random flipping and shifting with cutout) 
did not perform the best. One possible reason is that the generator sometimes also generates samples with augmented artifacts (e.g., cutout). If such artifacts do not exist in the real dataset, it might lead to worse FID performance.  

\section{Conclusion}
In this paper, we propose a simple, effective, and computationally cheap method -- consistency regularization -- to improve the performance of GANs.
Consistency regularization is compatible with spectral normalization and results in improvements in all of the many contexts in which we evaluated it.
Moreover, we have demonstrated consistency regularization is more effective than other regularization methods under different loss functions, neural architectures and optimizer hyper-parameter settings.
We have also shown simply applying consistency regularization on top of state-of-the-art GAN models can further greatly boost the performance. 
Finally, we have conducted a thorough study on the design choices and hyper-parameters of consistency regularization. 

\section*{Acknowledgments}
We thank Colin Raffel for feedback on drafts of this article.
We also thank Marvin Ritter, Michael Tschannen and Mario Lucic for answering our questions of using compare GAN codebase for large scale GAN evaluation. 

\bibliography{iclr2020_conference}
\bibliographystyle{iclr2020_conference}

\newpage
\section*{Appendix}

\appendix
\counterwithin{figure}{section}
\renewcommand{\thefigure}{A\arabic{figure}}
\setcounter{table}{0}
\renewcommand{\thetable}{A\arabic{table}}

\input{appendix}

\end{document}

%% file: math_commands.tex
\usepackage{amsmath,amsfonts,bm}

\def\eqref#1{equation~\ref{#1}}

\def\1{\bm{1}}

\DeclareMathAlphabet{\mathsfit}{\encodingdefault}{\sfdefault}{m}{sl}
\SetMathAlphabet{\mathsfit}{bold}{\encodingdefault}{\sfdefault}{bx}{n}

\newcommand{\normltwo}{L^2}


%% file: appendix.tex
\section{Hyperparameter settings of optimizer} \label{sec:hyper_settings}

\begin{table}[h]
\centering
\begin{tabular}{c|cccc}
 \hline
Setting   & $lr$ &  $\beta_1$ & $\beta_2$ & $N_{dis}$ \\ \cline{1-5}
 A  & 0.0001  & 0.5  & 0.9   &  5    \\
 B  & 0.0001  & 0.5  & 0.999 &  1    \\ 
 C  & 0.0002  & 0.5  & 0.999 &  1    \\
 D  & 0.0002  & 0.5  & 0.999 &  5    \\
 E  & 0.001   & 0.5  & 0.9   &  5    \\
 F  & 0.001   & 0.5  & 0.999 &  5     \\
 G  & 0.001   & 0.9  & 0.999 &  5     \\
 
 \hline \hline
\end{tabular}
\caption{Hyper-parameters of the optimizer used in our experiments. 
} 
\label{tab:different_adam_parameter}
\end{table}

 Here, similar as the experiments in \citet{Miyato18a, compare_gan}, we evaluate all regularization methods across 7 different hyperparameters settings for (1) learning rate $lr$  (2)  first and second order momentum parameters of Adam $\beta_1$, $\beta_2$ (3) number of the updates of the discriminator per generator update, $N_{dis}$. The details of all the settings are shown in Table \ref{tab:different_adam_parameter}. Among all these 7 settings, A-D are the "good" hyperparameters used in previous publications \citep{Radford15, WGANGP, compare_gan}; E, F, G are the "aggressive" hyperparameter settings introduced by \citet{Miyato18a} to test model performance under noticeably large learning rate or disruptively high momentum. In practice, we find setting C generally works the best for SNDCGAN and setting D is the optimal setting for ResNet. These two settings are also the default settings in the Compare GAN codebase for the corresponding network architectures. 
 
 \begin{figure}[hbt]
    \centering
    \includegraphics[width=140mm]{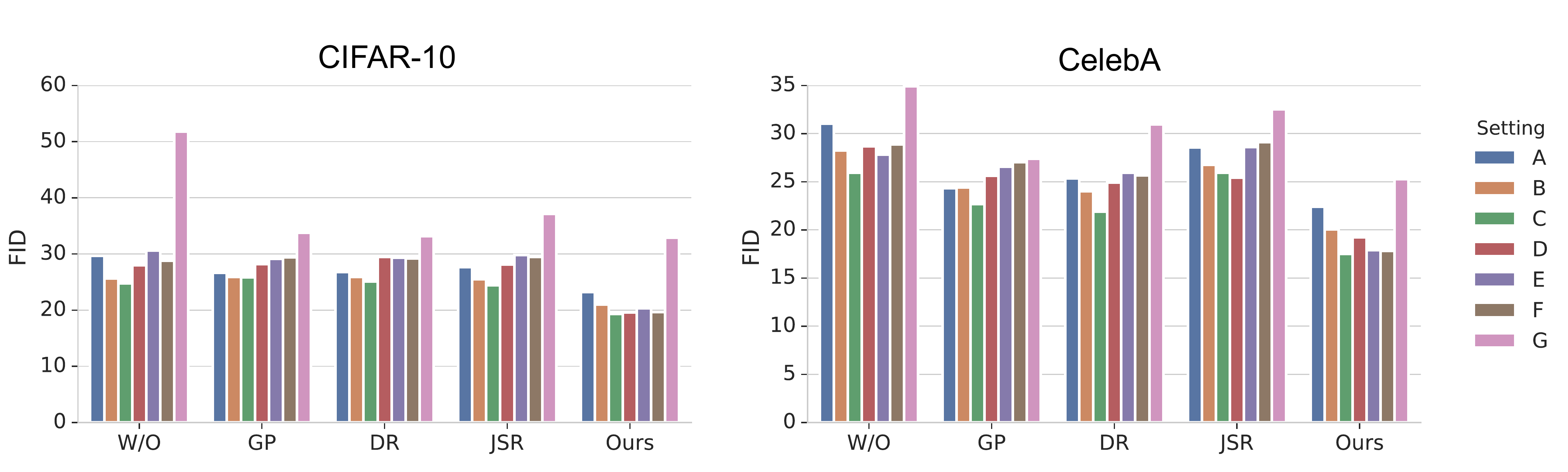}
    \caption{
    Comparison of FID scores with different optimizer settings.
    }
    \label{fig:optimizer_setting}
\end{figure}
 
Figure \ref{fig:optimizer_setting} displays the FID score of all methods with 7 settings A-G. We can observe that consistency regularization is fairly robust even for some of the aggressive hyperparameter settings. In general, the proposed consistency regularization can generate better samples with different optimizer settings compared with other regularization methods.

\newpage 
\renewcommand{\thefigure}{B\arabic{figure}}
\setcounter{figure}{0}

\section{Comparison of different regularization methods when spectral normalization is not used} \label{sec:no_sn}

\begin{figure}[hbt]
    \centering
    \includegraphics[width=140mm]{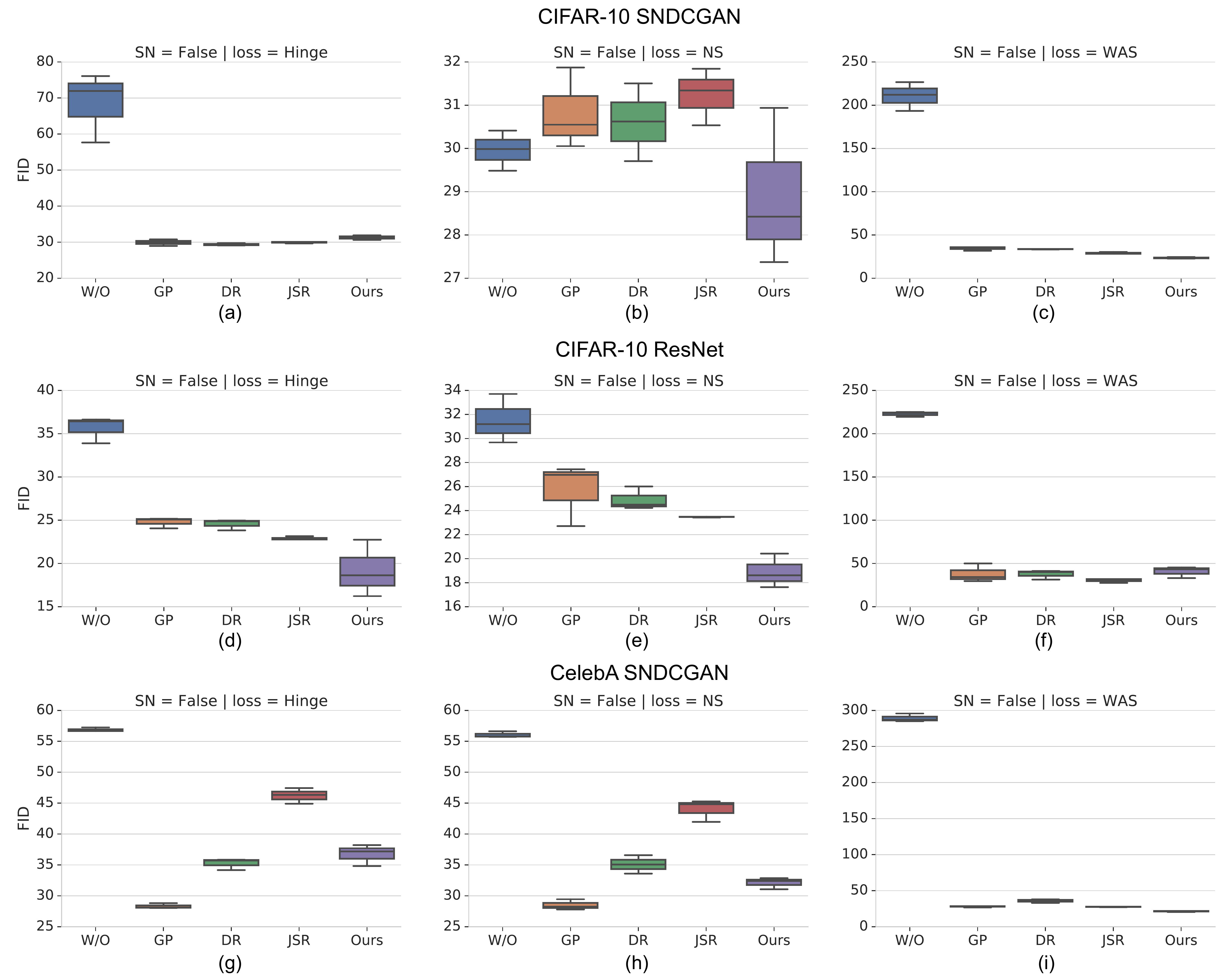}
    \caption{
    Comparison of FID scores when SN is not used.
    }
    \label{fig:sn_false_all}
\end{figure}

Here, we compare different regularization methods when spectral normalization (SN) is not used. As shown in Figure \ref{fig:sn_false_all}, our consistency regularization always improves the baseline model (W/O). It also achieves the best FID scores in most of the cases, which demonstrates that consistency regularization does not depend on spectral normalization. By comparing with the results in Figure \ref{fig:loss_fn} and Figure \ref{fig:resnet_loss}, we find adding spectral normalization will further boost the results. More importantly, the consistency regularization is only method that improve on top of spectral normalization without exception. The other regularization methods do not have this property. 

\section{Training Speed}
\renewcommand{\thefigure}{C\arabic{figure}}
\setcounter{figure}{0}
\renewcommand{\thetable}{C\arabic{table}}
\setcounter{table}{0}

Here we show the actual training speed of discriminator updates for SNDCGAN on CIFAR-10 with NVIDIA Tesla V100. Consistency regularization is around $1.7$ times faster than gradient based regularization techniques.

\begin{table}[hbt]
\centering
\begin{tabular}{l|ccccc}
 \hline
Method   & W/O &  GP & DR & JSR  & Ours (CR-GAN) \\ 
\hline
Speed (step/s)  & 66.3 & 29.7  & 29.8 & 29.2   & 51.7 \\

 \hline \hline
\end{tabular}
\caption{Training speed of discriminator updates for SNDCGAN on CIFAR-10. 
} 
\label{tab:speed_comparison}
\end{table}
\newpage 
\section{Generated samples for unconditional image generation}

\renewcommand{\thefigure}{D\arabic{figure}}
\setcounter{figure}{0}
\renewcommand{\thetable}{D\arabic{table}}
\setcounter{table}{0}

\begin{figure}[hbt]
    \centering
    \includegraphics[width=130mm]{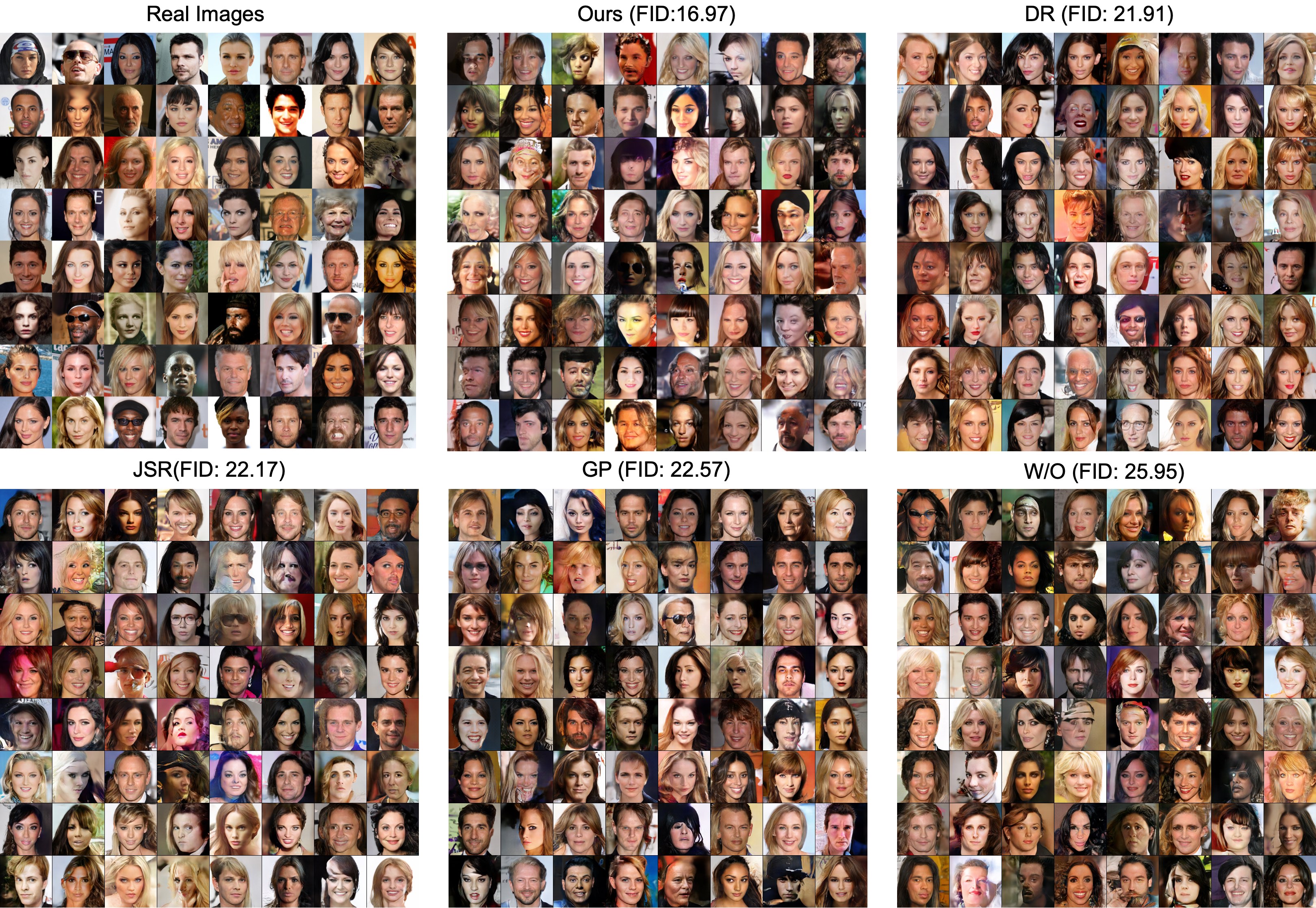}
    \caption{
    Comparison of generated samples of CelebA.
    }
    \label{fig:celeba_samples}
\end{figure}

\begin{figure}[hbt]
    \centering
    \includegraphics[width=130mm]{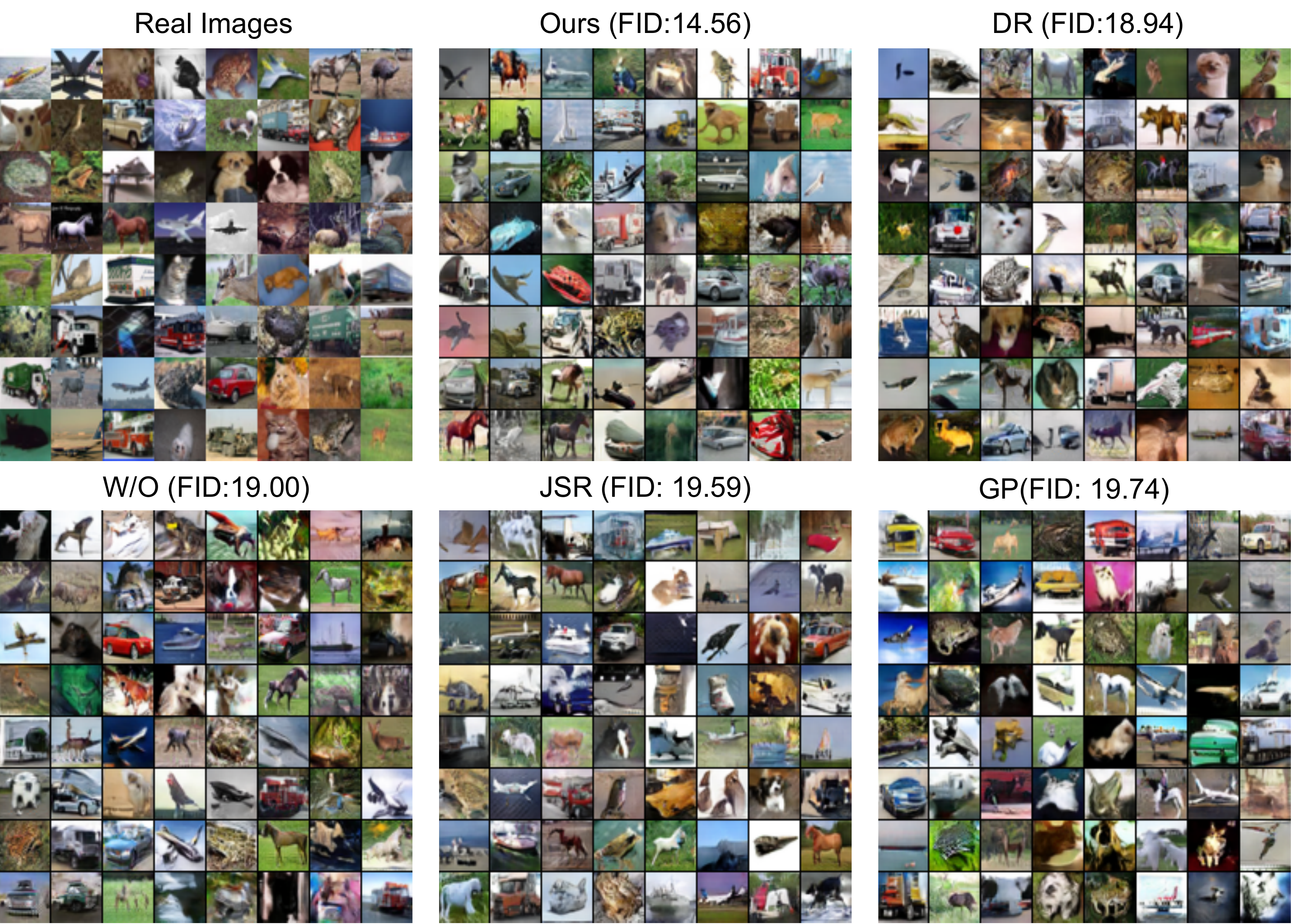}
    \caption{
    Comparison of generated samples for unconditional image generation on CIFAR-10 with a ResNet architecture. 
    }
    \label{fig:cifar_samples_unconditional}
\end{figure}

\begin{figure}[hbt]
    \centering
    \includegraphics[width=130mm]{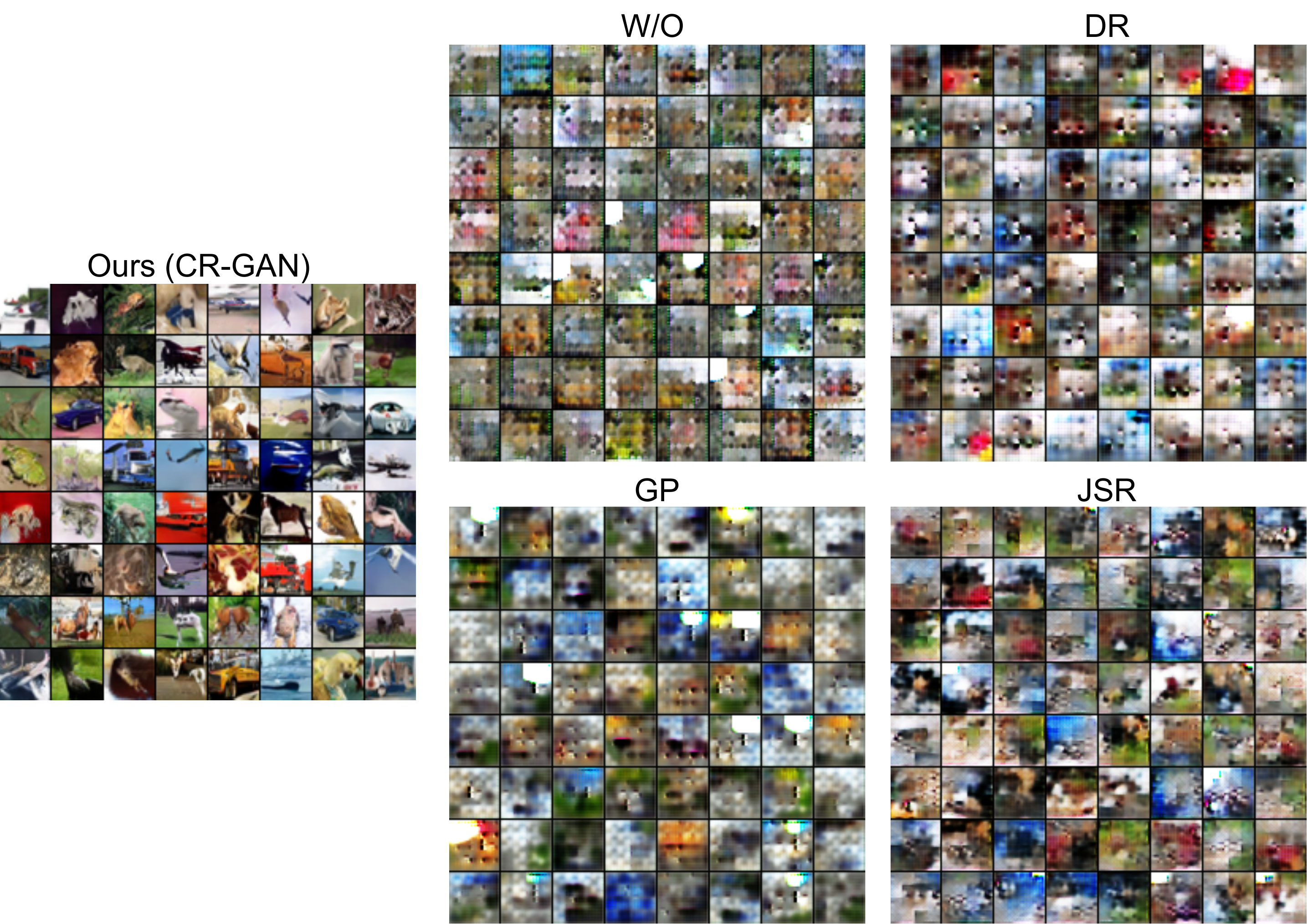}
    \caption{
    Comparison of unconditional generated samples on CIFAR-10 with a ResNet  architecture,  Wasserstein loss and spectral normalization. This is a hard hyperparameter setting where the baseline and previous regularization methods fail to generate reasonable samples. Consistency Regularization is the only regularization method that can generate satisfactory samples in this setting. FID scores are shown in sub-plot (c) of Figure \ref{fig:resnet_loss}.
    }
    \label{fig:stability_sample}
\end{figure}

\section{Generated samples for conditional image generation}
\renewcommand{\thefigure}{E\arabic{figure}}
\setcounter{figure}{0}
\renewcommand{\thetable}{E\arabic{table}}
\setcounter{table}{0}

\begin{figure}[hbt]
    \centering
    \includegraphics[width=140mm]{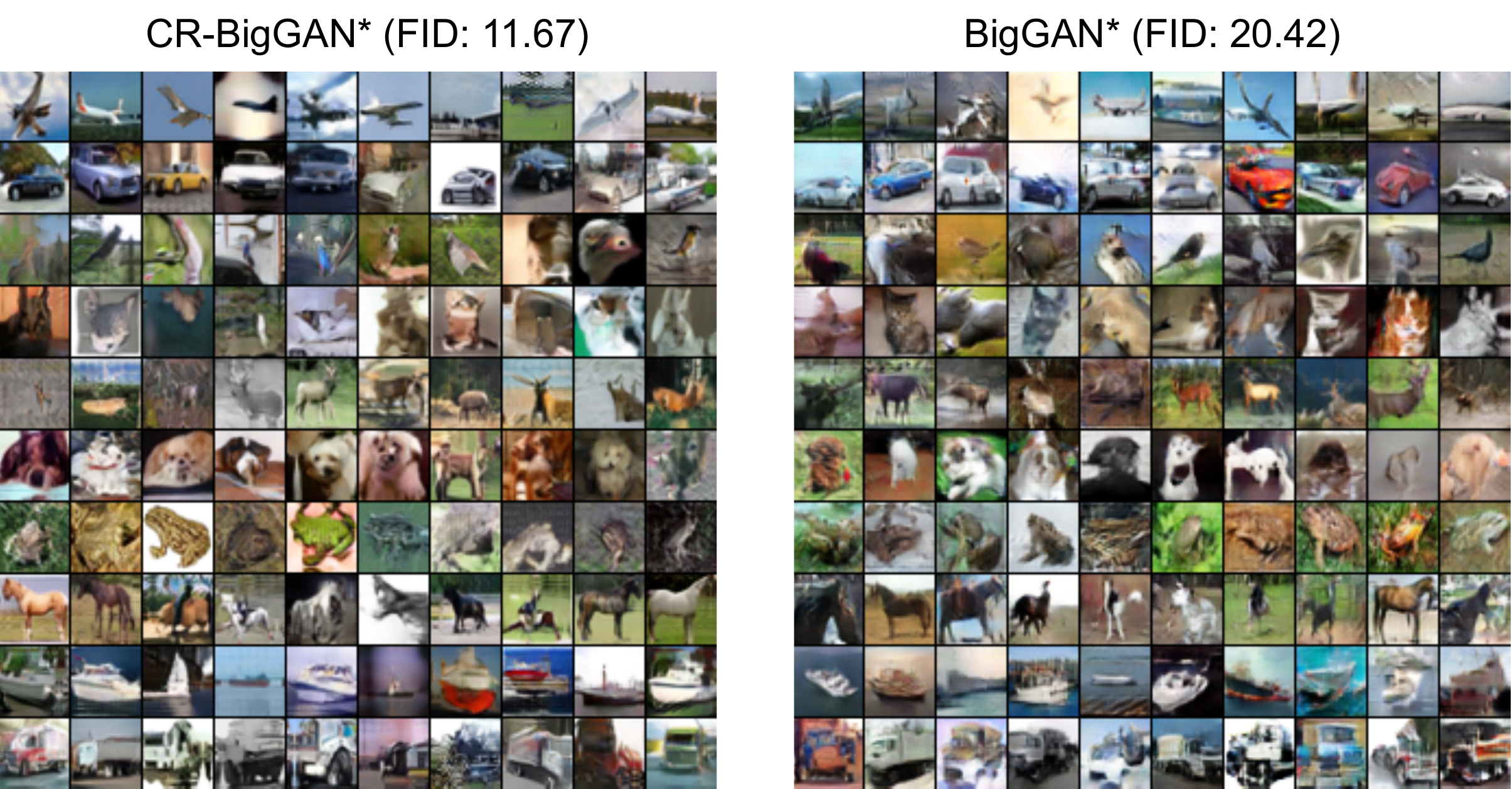}
    \caption{
    Comparison of generated samples for conditional image generation on CIFAR-10. Each row shows the generated samples of one class.
    }
    \label{fig:cifar_samples_conditional}
\end{figure}
\newpage
\begin{figure}[hbt!]
    \centering
    \includegraphics[width=138mm]{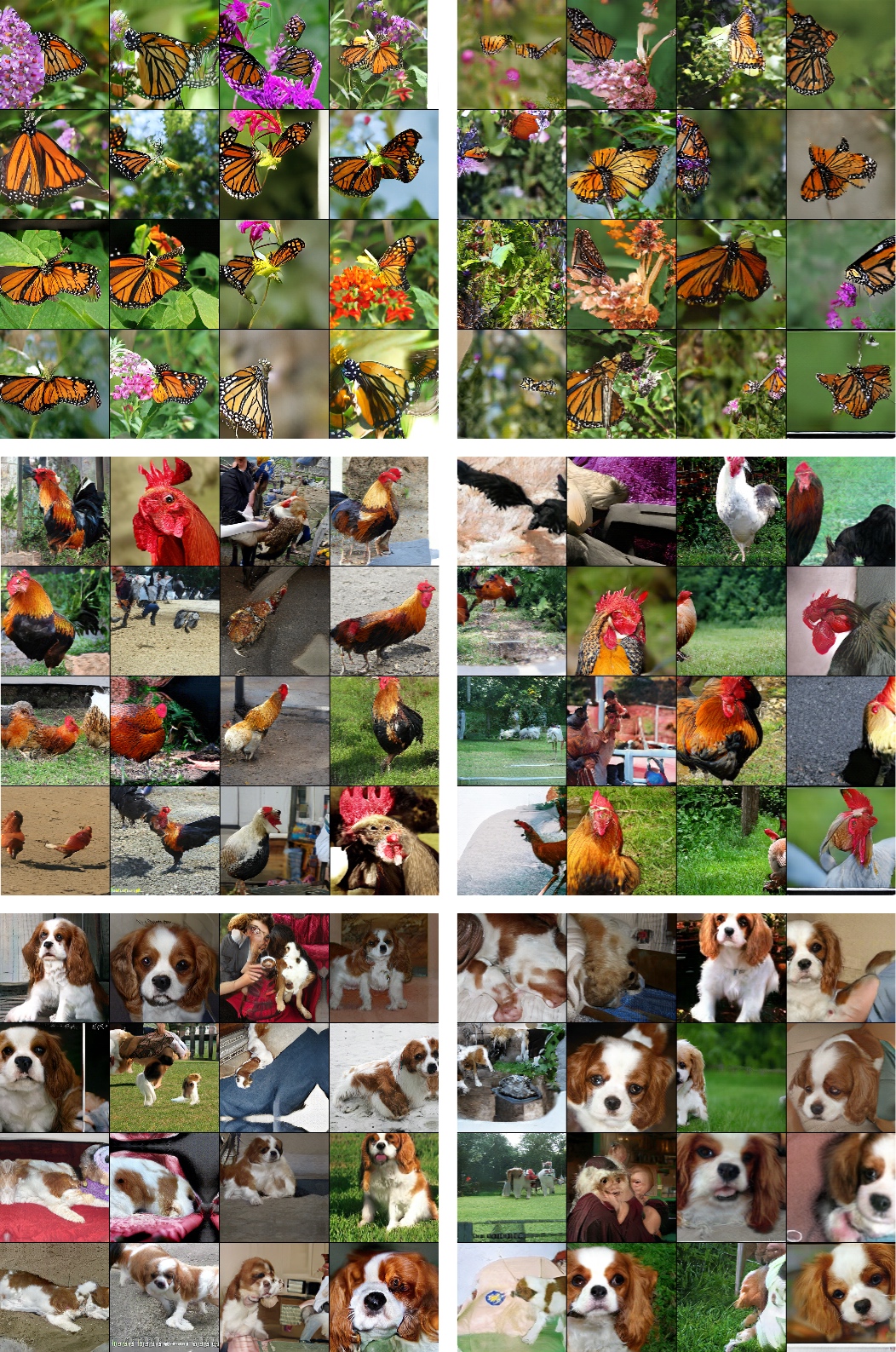}
    \caption{
        Comparison of conditionally generated samples of BigGAN* and CR-BigGAN* on ImageNet. (\textbf{Left}) Generated samples of CR-BigGAN*. (\textbf{Right}) Generated samples of BigGAN*. 
    }
    \label{fig:imagenet_ours}
\end{figure}
\begin{figure}[hbt!]
    \centering
    \includegraphics[width=138mm]{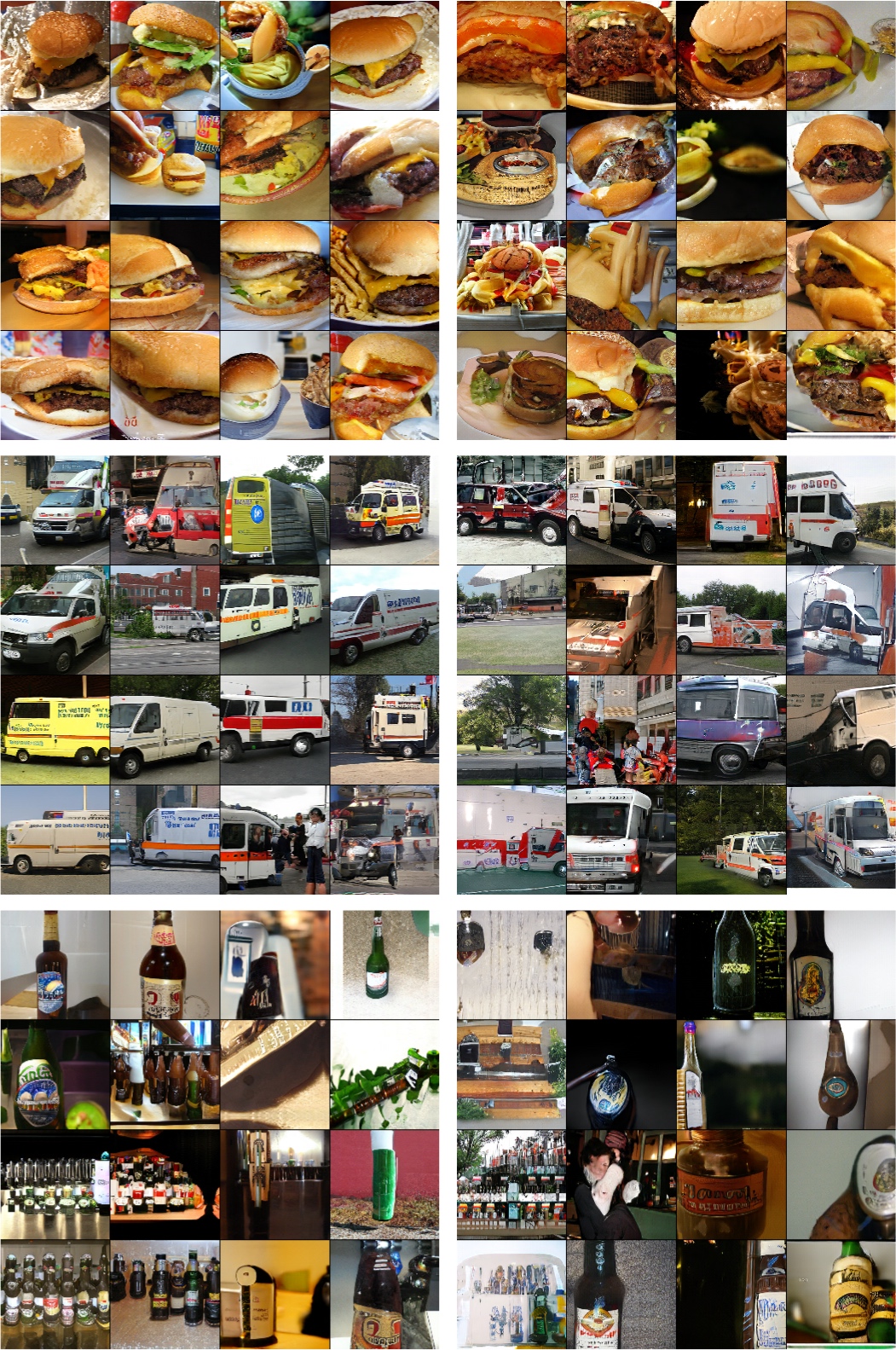}
    \caption{
    More results for conditionally generated samples of BigGAN* and CR-BigGAN* on ImageNet. (\textbf{Left}) Generated samples of CR-BigGAN*. (\textbf{Right}) Generated samples of BigGAN*. 
    }
    \label{fig:imagenet_baseline}
\end{figure}

\section {Comparison with inception score} \label{sec:is_section}
\renewcommand{\thefigure}{F\arabic{figure}}
\setcounter{figure}{0}
\renewcommand{\thetable}{F\arabic{table}}
\setcounter{table}{0}
Inception Score (IS) is another GAN evaluation metric introduced by \citet{salimans2016improved}. Here, we compare the Inception Score of the unconditional generated samples on CIFAR-10. As shown in Table \ref{tab:main_inception_comparision},  Figure \ref{fig:cifar_dcgan_is} and Figure \ref{fig:cifar_resnet_is}, consistency regularization achieves the best IS result with both SNDCGAN and ResNet architectures. 

\begin{table}[hbt]
\centering
\begin{tabular}{l|ccccc}
 \hline
Setting   & W/O &  GP & DR & JSR  & Ours (CR-GAN)\\ 
\hline
 CIFAR-10 (SNDCGAN)  & 7.54 & 7.54  & 7.54  & 7.52   & \textbf{7.93} \\
 CIFAR-10 (ResNet)  &8.20 & 8.04  & 8.09  & 8.03 &   \textbf{8.40}\\

 \hline \hline
\end{tabular}
\caption{Best Inception Score for unconditional image generation on CIFAR-10.
} 
\label{tab:main_inception_comparision}
\end{table}

\begin{figure}[hbt!]
    \centering
    \includegraphics[width=140mm]{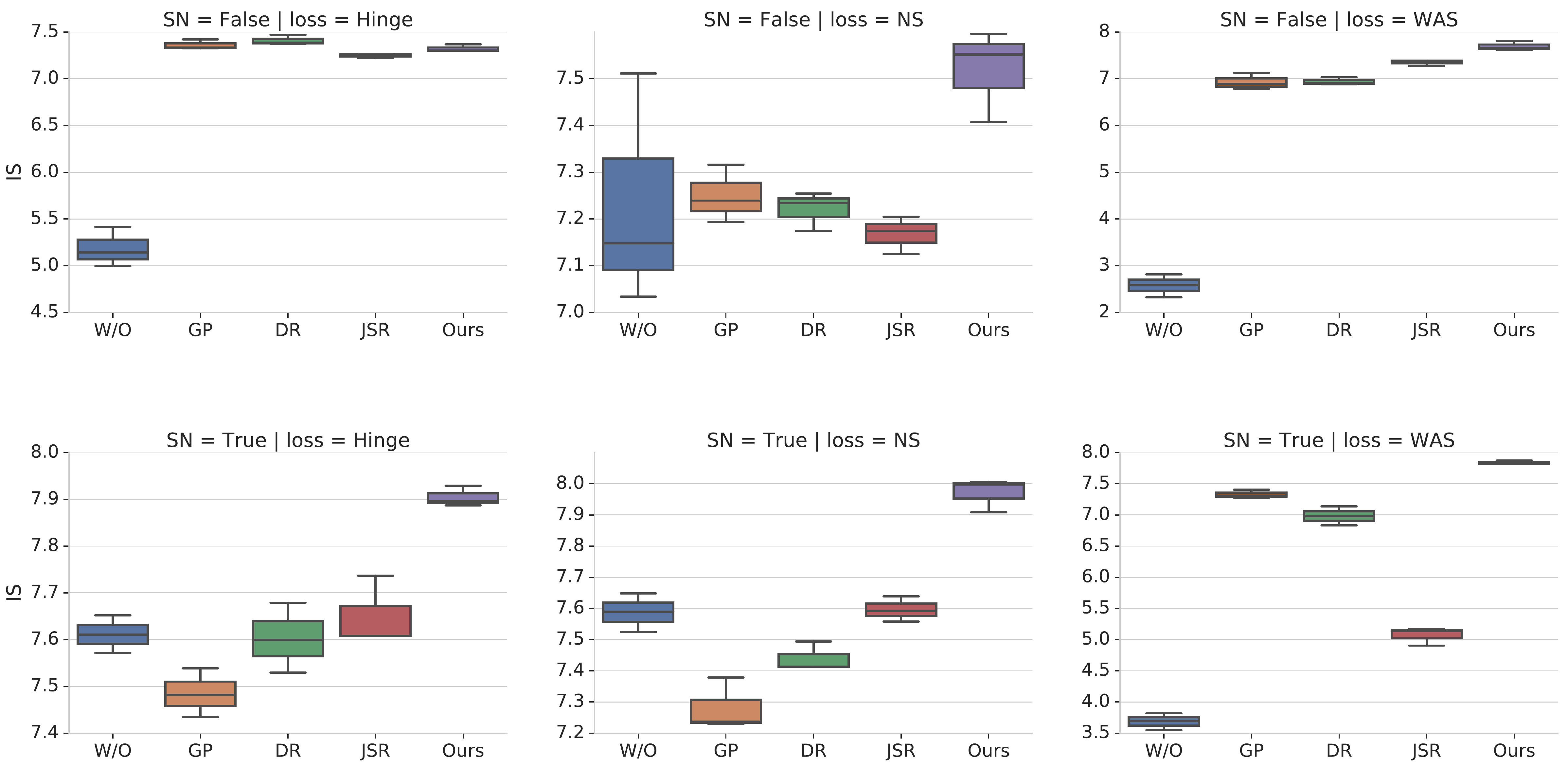}
    \caption{
    Comparison of IS with a SNDCGAN architecture on different loss settings. Models are trained on CIFAR-10.
    }
    \label{fig:cifar_dcgan_is}
\end{figure}

\begin{figure}[hbt!]
    \centering
    \includegraphics[width=140mm]{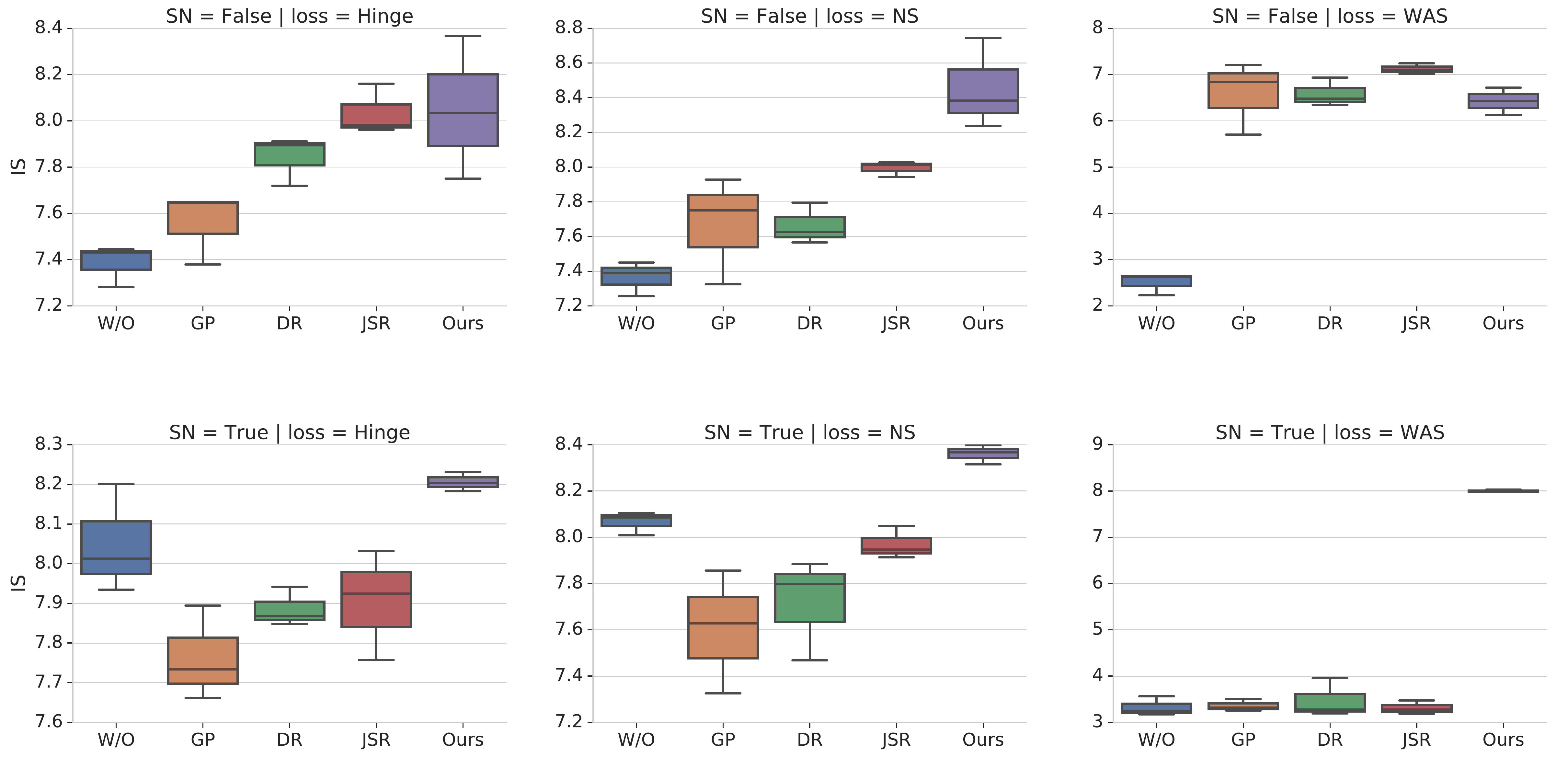}
    \caption{
    Comparison of IS with a ResNet architecture on different loss settings. Models are trained on CIFAR-10.
    }
    \label{fig:cifar_resnet_is}
\end{figure}

\section {Effect of the number of layers regularized in Discriminator } \label{sec:num_layers}

\renewcommand{\thefigure}{G\arabic{figure}}
\setcounter{figure}{0}
\renewcommand{\thetable}{G\arabic{table}}
\setcounter{table}{0}

\begin{figure}[hbt!]
    \centering
    \includegraphics[width=140mm]{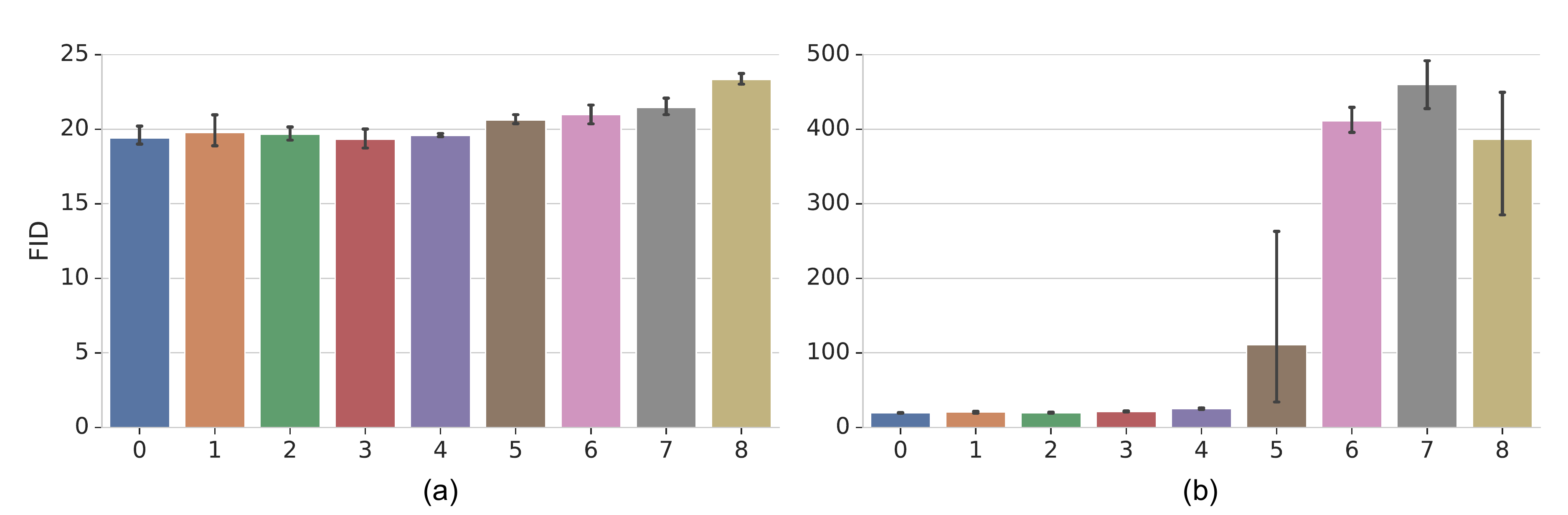}
    \caption{
    Comparison of consistency regularization on different number of intermediate layers: (a) first weight setting, where the weight for each layer is the inverse of its feature dimension (b) second weight setting, where each layer has equal weight. 
    }
    \label{fig:layer_sensitivity}
\end{figure}

Here, we examine the effect of the number of layers regularized in discriminator.
In this experiment, we use SNDCGAN architecture with NS loss on the CIFAR-10 dataset.
There are 8 intermediate layers in the discriminator.
To start, we add consistency only to the last layer (0 intermediate layers).
Then we gradually enforce consistency for more intermediate layers.
We use two weighting variations to combine the consistency loss across different layers.
In the first setting, the weight of each layer is the inverse of feature dimension $d_j$ in that layer,
which corresponds to $\lambda_j = \sfrac{1}{d_j}$ in Equation~\ref{eq:ori_eq}.
In the second setting, we give equal weight to each layer, which corresponds to  $\lambda_j=1$.
The results for both settings are shown in Figure \ref{fig:layer_sensitivity}.
In both settings, we observe that consistency regularization on the final layer achieves reasonably good results.
Adding the consistency to first few layers in the discriminator harms the performance.
For simplicity, we only add consistency regularization in the final layer of the discriminator in the rest of our experiments. 

\section {Consistency Regularization on the generated samples} \label{sec:generated_sample}

\renewcommand{\thefigure}{H\arabic{figure}}
\setcounter{figure}{0}
\renewcommand{\thetable}{H\arabic{table}}
\setcounter{table}{0}

\begin{figure}[hbt!]
    \centering
    \includegraphics[width=0.49\linewidth]{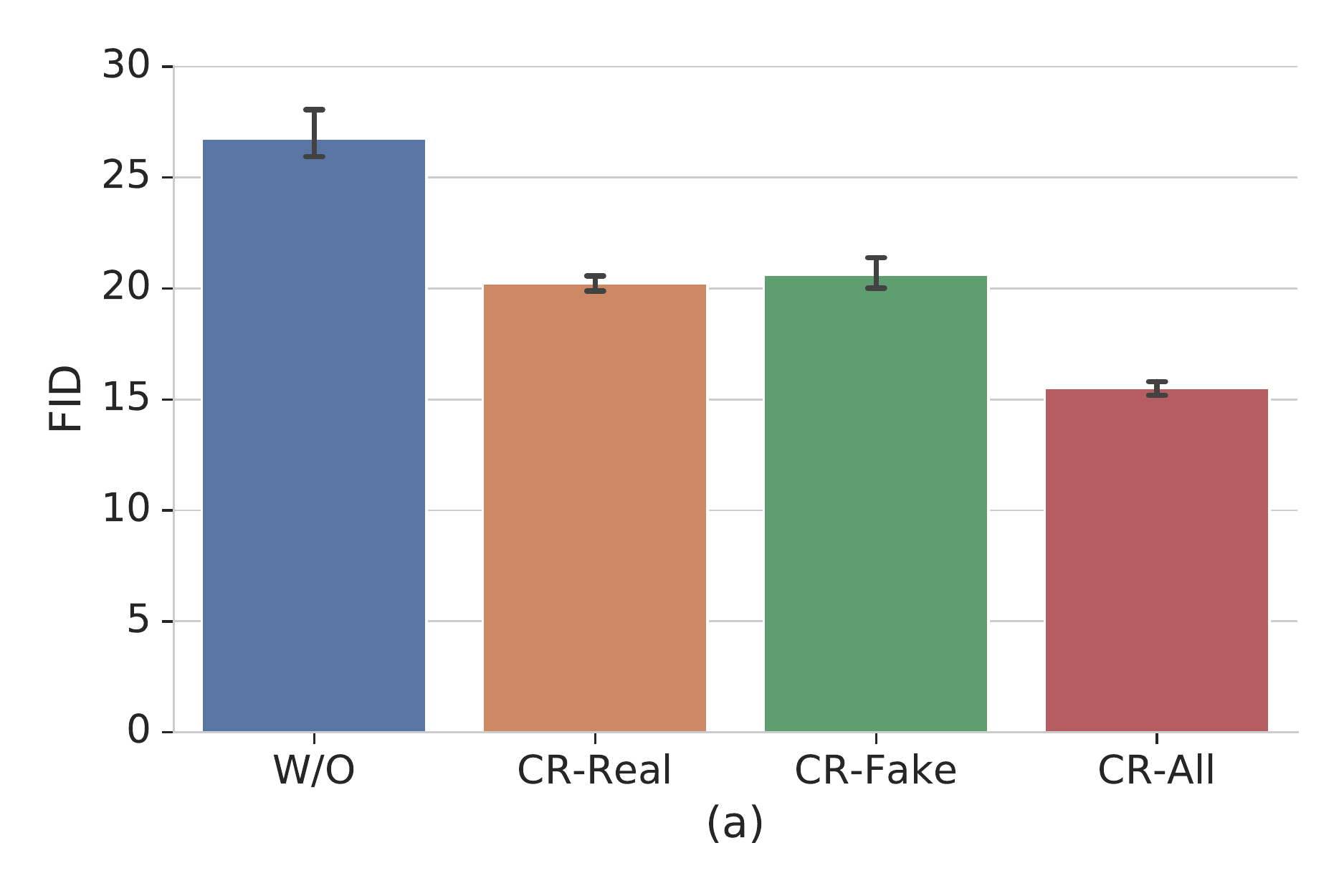}
    \includegraphics[width=0.49\linewidth]{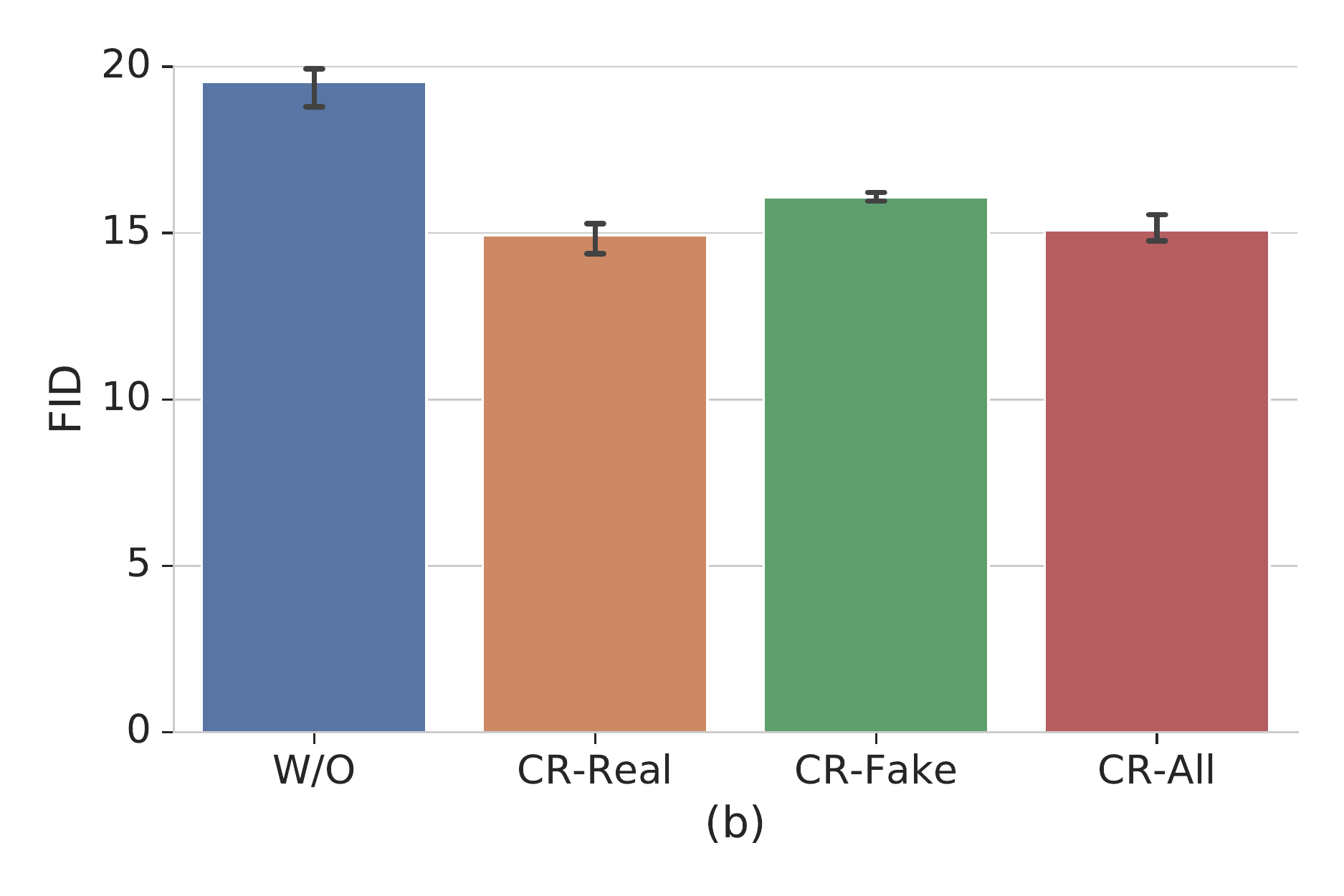}
    \includegraphics[width=0.49\linewidth]{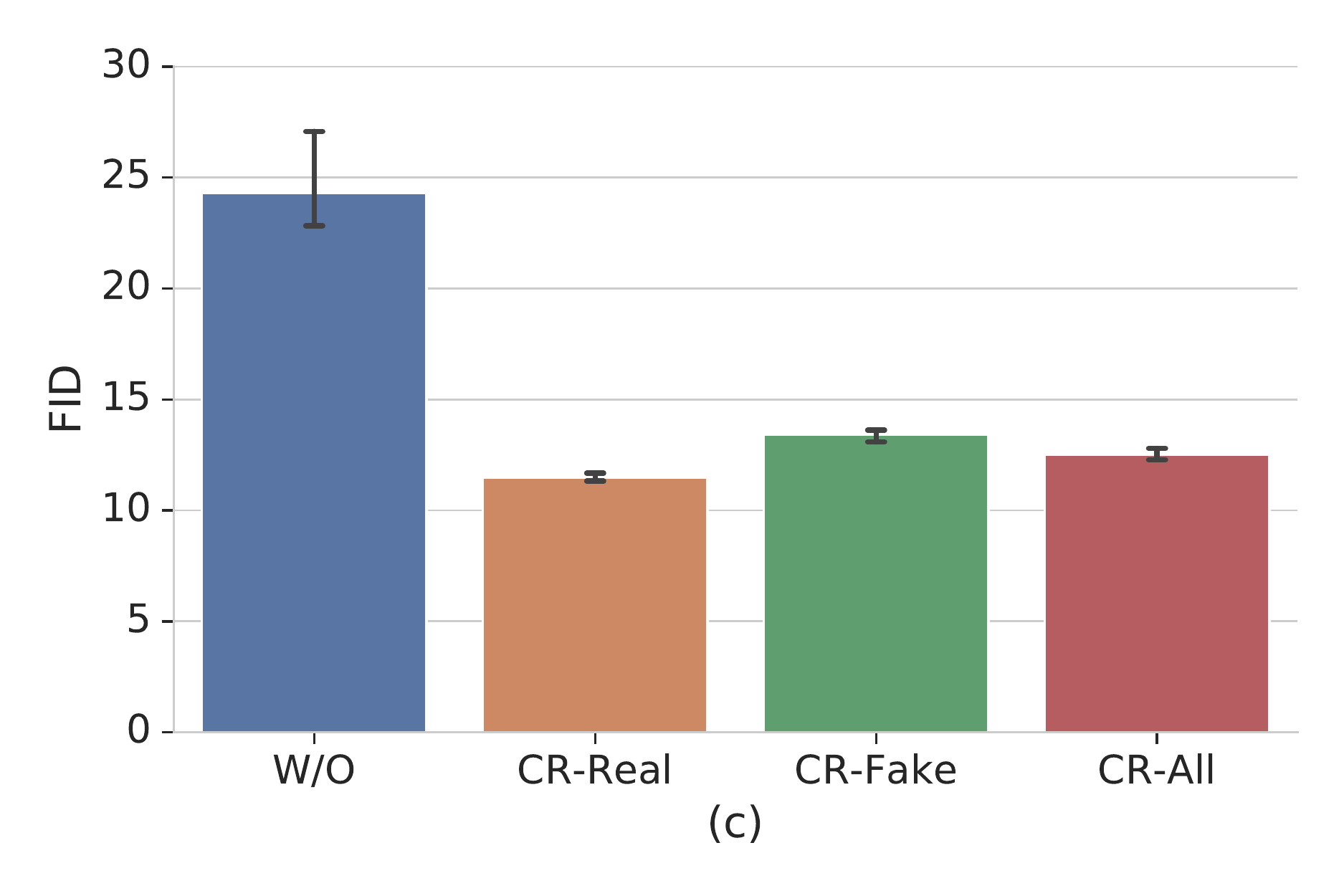}
    
    \caption{
    Comparison of FID scores with no consistency regularization (W/O), regularization only on the real samples (CR-Real), consistency regularization only on the fake samples produced by the generator (CR-Fake) and regularization on both real and fake samples (CR-All) for (a) unconditional image generation on CIFAR-10 with SNDCGAN, (b) unconditional image generation on CIFAR-10 with ResNet, (c) conditional image generation on CIFAR-10 with CR-BigGAN*.
    }
    \label{fig:cr_fake}
\end{figure}

In this section, we investigate the effect of adding consistency regularization for the generated samples.
We compare four settings, no consistency regularization (W/O), regularization only on the real samples (CR-Real),
consistency regularization only on the fake samples produced by the generator (CR-Fake) and regularization on both real and fake samples (CR-All).
CR-Real is presented in Algorithm \ref{alg:main}. 
CR-Fake has similar computational cost as CR-Real and CR-All doubles the computational cost, 
since both the augmented real and fakes samples need to be fed into the discriminator to calculate the consistency loss.
As shown in Figure \ref{fig:cr_fake}, CR-Real, CR-Fake and CR-All are always better than the baseline without consistency regularization.
In addition, CR-Real is consistently better than CR-Fake.
It is interesting to note that CR-All is not always better than CR-real given the extra computational costs and stronger regularization.
For example, CR-All improves FID from 20.21 of CR-Real to 15.51 for SNDCGAN, but it also gives slightly worse results for ResNet  (14.93 vs 15.07)
and for CR-BigGAN* (11.48 vs 12.51).
We observe that enforcing additional consistency on the generated samples gives more performance gain when the model capacity is small and that gain
decreases when model capacity increases.
For computational efficiency and simplicity of the training algorithm, we use consistency regularization on real samples for the rest of our 
experiments.

